# The Code2Text Challenge: Text Generation in Source Code Libraries

**Kyle Richardson**[†], **Sina Zarrieß**[‡] and **Jonas Kuhn**[†]

[†]Institute for Natural Language Processing, University of Stuttgart, Germany
`kyle@ims.uni-stuttgart.de`
[‡]Dialogue Systems Group // CITEC, Bielefeld University, Germany
`sina.zarriess@uni-bielefeld.de`

## Abstract

We propose a new shared task for tactical data-to-text generation in the domain of source code libraries. Specifically, we focus on text generation of function descriptions from example software projects. Data is drawn from existing resources used for studying the related problem of semantic parser induction (Richardson and Kuhn, 2017b; Richardson and Kuhn, 2017a), and spans a wide variety of both natural languages and programming languages. In this paper, we describe these existing resources, which will serve as training and development data for the task, and discuss plans for building new independent test sets.

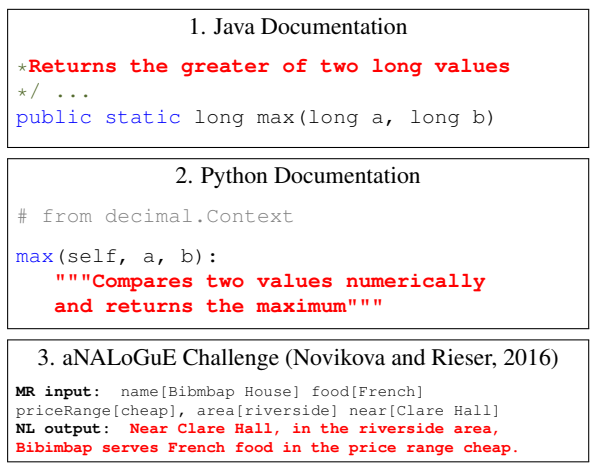

**Figure 1:** Example source code documentation, or *docstrings* in 1-2, and an example MR/text pair.

## 1 Introduction

Source code libraries are collections of computer programs/instructions expressed in a target programming language that aim to solve some set of problems. Within these libraries, the designers of the code often use natural language to describe how various internal components work. For example, Figure 1.1 shows a docstring description (in red) for the `max` function in the Java standard library, which explains what the function does (i.e., *Returns the greater of*), and the types of arguments that the function takes (i.e., *two long values*). Similarly, a related function and its documentation for the Python programming language is shown in Figure 1.2.

Given the tight coupling between such high-level text and lower-level representations, automatically extracting parallel datasets in this domain, consisting of short text descriptions and code templates, is rather straightforward. Such datasets can then be used to study various translation problems, including translating text to code templates (i.e., semantic parsing), or generating representations to text (i.e., data-to-text generation). In previous work (Richardson and Kuhn, 2017b), we looked at using source code libraries to study the first problem, and have collected datasets for 43 software libraries across 7 natural languages. We have also created a tool, called `Function Assistant`, for extracting new datasets from arbitrary software projects (Richardson and Kuhn, 2017a).

In this paper, we propose using these resources for studying the second problem. The task can be described as follows: given a source code library dataset or collection of datasets consisting of text and code template pairs, e.g., the text *Returns the greater of two..* and the function representation `static long max(..)` in Figure 1.1, create a model that generates well-formed natural language descriptions of these formal code inputs. This task involves solving several sub-tasks, chief among them being *lexicalization*, or the problem of how to verbalize the function name and return value (here using a VP *returns the greater of*), and the function's arguments (expressed here as a plural expression, *two long values*). As shown in Figure 1.3, existing lexicalization tasks tend to involve input representations that have considerable lexical overlap with the verbal output, which is not the case with our datasets and therefore makes our problem more difficult. In addition, there is the problem of *realization*, or here aggregating the description as a sentence with an implied subject containing a transitive verb and a complex object, where the referring expression is attached as a PP.

The available resources make the task highly multilingual, both in terms of the input programming languages and output natural languages. Since programming languages differ in terms of representation conventions, each formal language provides unique challenges related to these differences. For example, statically-typed languages, such as Java in Figure 1.1., contain more information about argument and return values than dynamically-typed languages, such as Python in Figure 1.2.

In what follows, we motivate this task by discussing related work. We also discuss the current datasets and plans to build new independent test sets.

## 2 Related Work

*Data-to-text* generation concerns the problem of generating well-formed, natural language descriptions from non-linguistic, formal meaning representations (Gatt and Krahmer, 2017). In our case, the input to a given generation system is a source code representation. In order to learn a natural language generation (NLG) system from data, a parallel corpus containing pairs of inputs and outputs must be constructed. In many studies on data-to-text generation, these parallel resources are relatively small, cf. work on sportscasting (Chen and Mooney, 2008), weather reporting (Belz, 2008; Liang et al., 2009), or biology facts (Banik et al., 2013). We follow similar efforts to build automatic parallel resources (Belz and Kow, 2010) by mining example software libraries for (raw) pairs of short text descriptions and function representations.

A recent trend is the use of crowd-sourcing to obtain parallel NLG data (Wen et al., 2016; Novikova et al., 2016; Gardent et al., 2017). Crowd-workers are presented with some meaning representation (MR, e.g., triples from a knowledge base) and asked to verbalize these representations in natural language. For example, the **MR input** in Figure 1.3 in the restaurant domain is verbalized as the **NL output** text. While this method allows for fast annotation, and thus solves the data scarcity problem, it also raises some new issues. For instance, sentences or utterances are produced by crowd-workers without much context, which puts to question the naturalness of the resulting text. Novikova et al. (2016) compare collecting data from logic-based MRs, of the type shown in Figure 1.3, and pictorial MRs, and find that the former approach leads to less natural and less informative descriptions. This seems to be related to the problem that the natural language sentence is a very close verbalization of the "logic" input, i.e., many terms in the MR can be simply taken up in the sentence.

Our approach relies on naturally occurring verbal descriptions produced by human developers. Our input data (source code representations) seems more abstract than previously used representations e.g. in the restaurant domain where many lexical items in the target utterance already appear in the MR. Thus, our input data is more "naturally occurring" in the sense that it has not been designed specifically for an NLG system (as compared to Wen et al. (2016) who randomly generate input representations) yet it still corresponds to a formal language. We expect that there is relatively little lexical correspondence between source code representations and verbal descriptions and that this is an interesting challenge for data-driven NLG, as simple "alignment" methods might fail to predict lexicalization.

While natural language generation in technical domains has long been of interest to the NLG community (Reiter et al., 1995), there has been renewed interest in this and other closely related topics over the last few years in NLP (Allamanis et al., 2015; Iyer et al., 2016; Yin and Neubig, 2017), making a shared task on the topic rather timely. While preparing the final version of this paper, we learned about the work of (Miceli Barone and Sennrich, 2017), who similarly look at generating text from automatically mined Python projects, using a similar set of tools as ours. This interest seems largely related to the wider availability of new data resources in the technical domain, especially through technical websites such as Github and StackOverflow. Rather than focus on unconstrained source code representations, as done in some of these studies, we believe that limiting the expressivity of the generation input to function representations within *known* software libraries allows for more controlled experimentation.

On the resource side, our datasets are taken from (Richardson and Kuhn, 2017b; Richardson and Kuhn, 2017a). These resources have been used to study the problem of semantic parser induction, which is the inverse of the proposed data-to-text task. Given the close connection between the two tasks, there is often considerable overlap between the techniques used to solve either problem, techniques that are largely drawn from work on statistical machine translation (Wong and Mooney, 2006; Belz and Kow, 2009) and parsing (Zettlemoyer and Collins, 2012; Konstas and Lapata, 2012). While some approaches to generation explicitly use semantic parsing methods (Wong and Mooney, 2007; Zarrieß and Richardson, 2013), a more systematic investigation into the relation between these two tasks seems missing, which is a topic that we hope to address in this shared task.

## 3 Task Description

Given a collection of datasets consisting of text $x$ and function representation $z$ pairs, or $D = \{(x,z)_i\}_i^n$, the goal is to create a generation system that can produce well-formed, natural language descriptions from these representations, or $\texttt{gen} : z \to x$. As discussed above, such descriptions should not only cover what the associated function does in general, but should also describe the function's various parameters. As a secondary (optional) task, we will allow generation systems that accommodate processing in the other direction to compete on the task of semantic parsing, $\texttt{sp} : x \to z$, or generating function representations from text input.

### 3.1 Main Research Questions

Recent data-driven approaches in NLG have been successful in modeling end-to-end generation from unaligned input-output, cf. (Angeli et al., 2010; Mairesse and Young, 2014; Dušek and Jurcicek, 2015; Wen et al., 2016). However, these system have been mostly tested on datasets (e.g., in the restaurant domain) that require describing very similar entities, entities that are encoded in MRs that have considerable lexical overlap with the target text output. A central research question is whether these end-to-end approaches scale to NLG settings that involve substantially harder lexicalization problems, such as with our datasets where the overlap is considerably less. Similarly, generating source code documentation also involves describing a wide variety of functions from many different libraries, meaning that many more lexical concepts need to be learned.

A more general question is the following: to what extent can one build a function to text generation system by relying only on example input-output pairs? This question is partly about the sufficiency of function representations for natural language generation, namely, are these representations detailed enough to serve as a reasonable knowledge representation for natural language? If not, what is missing? How do hybrid approaches, perhaps approaches that rely on linguistically well-founded translation constraints and information about natural language syntax, fare against purely data-driven systems that rely solely on input-output as evidence?

Finally, the semantic parsing task addresses the following questions: what is the precise relationship between semantic parsing and data-to-text generation? Does an improvement in one task lead to an improvement in the other task? Is data-to-text generation simply an *inverse* semantic parsing task (Gatt and Krahmer, 2017) or are the two tasks fundamentally different?

## 4 Datasets

### 4.1 Train and Development Sets

Figure 2 shows information about the two datasets that will be available for model development, which were first introduced in (Richardson and Kuhn, 2017b) and (Richardson and Kuhn, 2017a) respectively[1]. None of these datasets have been previously used for data-to-text generation. The first dataset consists of the standard library documentation for 9 programming languages, including Java, Ruby, PHP,

---
[1] Please see the original papers to get more detailed information about each dataset

| Dataset | # Software Projects | # Training Pairs | # Programming Languages | # Natural Languages |
|---|---|---|---|---|
| Standard Library Docs | 16 | 38,652 | 9 | 7 |
| Python 27 | 27 | 37,567 | 1 | 1 |

**Figure 2:** A description of the currently available software datasets for model development.

Python, Elisp, Haskell, Clojure, C, and Scheme. In addition, this resource includes documentation in 7 natural languages, including English, French, Spanish, Russian, Japanese, Turkish, and German. The second resource includes 27 publicly available Python projects, taken from the well-known *awesome Python* list of (Chen, 2017).

Each individual standard library documentation set or Python project consists of short text descriptions with function representations. While each function representation typically has only a *single* text description, background information in the documentation allows one to find related functions, and therefore related descriptions, which can be taken into account for training and evaluation. We note that there is wide variability in the size of each individual dataset, and some datasets are low-resource. One interesting research question is whether it is feasible to build a NLG system in these low-resource scenarios, and whether training on multiple languages can help. In our previous experiments, we built individual models for each parallel dataset, though participants will be free to build models that are trained on multiple projects if desired.

We also note that these datasets are constructed automatically, and our existing extraction tool does not do extensive text preprocessing. The motivation for this is that we can quickly construct new resources for model development and evaluation, though the result is that some of available text descriptions are noisy. We however regard this "noisiness" as an interesting technical challenge, and contrasts with other shared tasks where more carefully curated data is assumed.

### 4.2 Test Sets

The publicly available test sets will be used for evaluation (see details of the evaluation below). In order to ensure that participants are not fitting their models to these sets, we are proposing to build three additional evaluation sets, each corresponding to a different programming language. These resources will be built using the `Function Assistant` toolkit (Richardson and Kuhn, 2017a), which already supports building parallel datasets from arbitrary Python source code projects, and will soon have functionality for the Java language.

The first two test sets, or *evaluation tracks*, will be specific to the Python and Java language, and will consist of unseen function representation-text pairs for each language. By having two separate sets according to language, we can see whether generation quality differs between different types of programming languages. Taking an idea from the recent CoNLL 2017 shared task on dependency parsing, the third evaluation track will include examples from a *surprise* programming language that has not been observed during the training phase. The idea is to see how generation systems generalize to unobserved languages where the inputs vary slightly.

## 5 Evaluation, Baselines and Scheduling

Following other data-to-text shared tasks (Banik et al., 2013; Colin et al., 2016) and previous work on text generating from code (Iyer et al., 2016), we will use automatic evaluation metrics such as BLEU and METEOR to evaluate system output. We will also perform fluency-based human evaluation on a subset of each test set using student volunteers from the Institute for Natural Language Processing (IMS), at the University of Stuttgart, Germany.

To establish baseline results, we have already started a pilot study that uses phrase-based SMT to do generation. Such models have previously been used to establish strong baseline generation results (Belz and Kow, 2009; Wong and Mooney, 2007), and have the advantage of being easy to run using known open-source tools. Since these models only require parallel data, they also show what a purely input-output driven model is capable of achieving on these datasets.

All publicly available datasets are immediately available for system development. The goal is to develop the new test sets before the end of 2017, and for the evaluation to be carried out in summer 2018.

# Acknowledgement

This work and the current data collection effort is funded by the German Research Foundation (DFG) via SFB 732, project D2, at the University of Stuttgart. In addition, we acknowledge support by the Cluster of Excellence "Cognitive Interaction Technology" (CITEC; EXC 277) at Bielefeld University, which is also funded by the DFG.